\documentclass[11pt,a4paper]{article}
\usepackage[hyperref]{acl2020}
\usepackage{times}
\usepackage{verbatim}
\usepackage{amssymb}
\usepackage{latexsym}
\usepackage{amsmath}
\usepackage{dsfont}
\usepackage{graphicx}
\usepackage{textcomp}
\usepackage{booktabs}
\usepackage{dirtytalk}
\usepackage{amssymb,amsmath, graphicx}
\usepackage{tikz}
\usepackage{url}
\usepackage{subcaption}

\newcommand{\sqdiamond}[1][fill=black]{\tikz [x=1.2ex,y=1.85ex,line width=.1ex,line join=round, yshift=-0.285ex] \draw  [#1]  (0,.5) -- (.5,1) -- (1,.5) -- (.5,0) -- (0,.5) -- cycle;}%

\newcommand{\MyDiamond}[1][fill=black]{\mathop{\raisebox{-0.275ex}{$\sqdiamond[#1]$}}}

\DeclareMathOperator*{\minimize}{minimize}

\DeclareMathOperator*{\gen}{gen}

\DeclareMathOperator*{\enc}{enc}
\DeclareMathOperator*{\mone}{supp}
\DeclareMathOperator*{\rat}{ext}
\DeclareMathOperator*{\mtwo}{pred}
\usepackage{url}

\aclfinalcopy 
\def\aclpaperid{2315}

\newcommand{\methodname}[0]{\textsc{FRESH}}
\newcommand{\methodfullname}[0]{Faithful Rationale  Extraction from Saliency tHresholding}

\newcommand{\todo}[1]{\textcolor{red}{\small [TODO: #1]}}

\usepackage[shortlabels]{enumitem}
\usepackage{multirow}

\title{Learning to Faithfully Rationalize by Construction}

\author{Sarthak Jain \\
 Khoury College of Computer Sciences \\
  Northeastern University \\
  \texttt{jain.sar@northeastern.edu} \\\And
  Sarah Wiegreffe \\
  School of Interactive Computing \\
  Georgia Institute of Technology \\
  \texttt{saw@gatech.edu} \\\AND
  Yuval Pinter \\
  School of Interactive Computing \\
  Georgia Institute of Technology \\
  \texttt{uvp@gatech.edu} \\\And
  Byron C. Wallace \\
  Khoury College of Computer Sciences \\
  Northeastern University \\
  \texttt{b.wallace@northeastern.edu}}

\date{}

\begin{document}
\maketitle
\begin{abstract}

In many settings it is important for one to be able to understand \emph{why} a model made a particular prediction. In NLP this often entails extracting snippets of an input text `responsible for' corresponding model output; when such a snippet comprises tokens that indeed informed the model's prediction, it is a \emph{faithful} explanation. In some settings, faithfulness may be critical to ensure transparency. \newcite{lei2016rationalizing} proposed a model to produce faithful rationales for neural text classification by defining independent snippet extraction and prediction modules. However, the discrete selection over input tokens performed by this method complicates training, leading to high variance and requiring careful hyperparameter tuning. We propose a simpler variant of this approach that provides faithful explanations by construction. In our scheme, named \methodname{}, arbitrary feature importance scores (e.g.,~gradients from a trained model) are used to induce binary labels over token inputs, which an extractor can be trained to predict. An independent classifier module is then trained exclusively on snippets provided by the extractor; these snippets thus constitute faithful explanations, even if the classifier is arbitrarily complex. In both automatic and manual evaluations we find that variants of this simple framework yield predictive performance superior to `end-to-end' approaches, while being more general and easier to train.\footnote{Code is available at \url{https://github.com/successar/FRESH}}
\end{abstract}

\section{Introduction}
\label{section:intro}

\begin{figure*}[t]
    \centering
    \includegraphics[width=\linewidth]{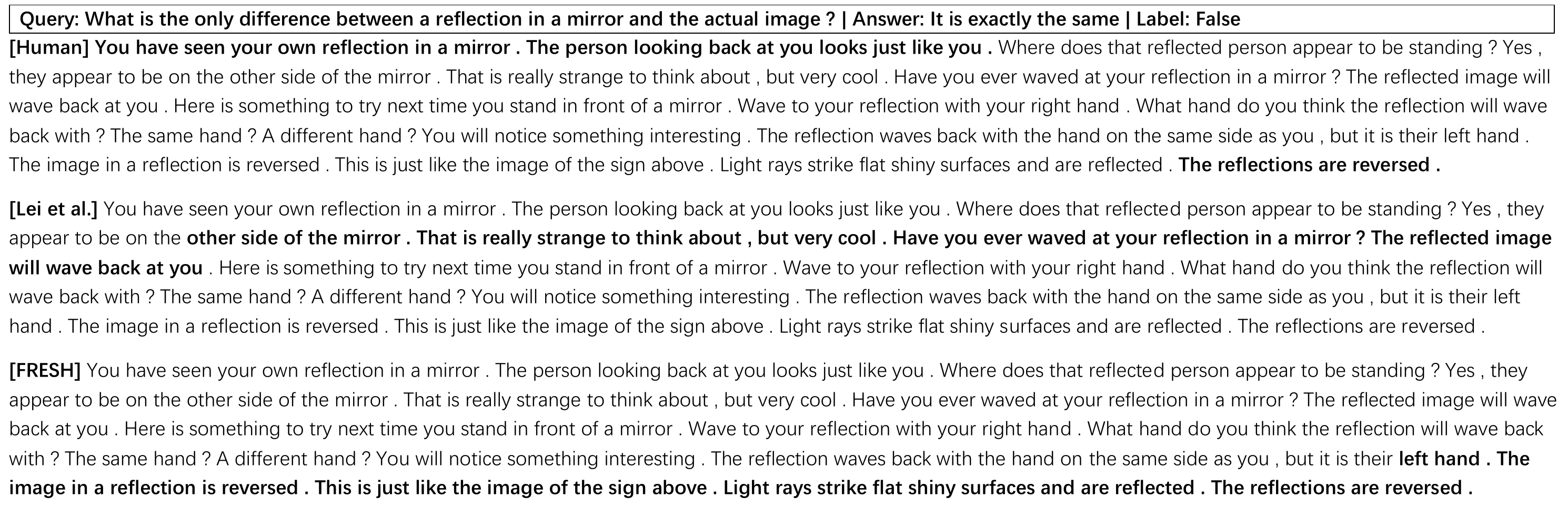}
    \caption{Contiguous rationales extracted using \citet{lei2016rationalizing} and \methodname{} models for an example from the MultiRC dataset. We also show the reference rationale associated with this example (top).}
    \label{fig:example}
\end{figure*}

Neural models dominate NLP these days, but it remains difficult to know \emph{why} such models make specific predictions for sequential text inputs. 
This problem has been exacerbated by the adoption of deep \emph{contextualized} word representations, whose architectures permit arbitrary and interdependent interactions between all inputs, making it particularly difficult to know which inputs contributed to any specific prediction.

Concretely, in a bidirectional RNN or Transformer model, the \emph{contextual embedding} for a word at position $j$ in instance $x$ may encode information from any or all of the tokens at positions $1$ to $j$-1 and $j$+1 to $|x|$.
Consequently, continuous scores such as attention weights~\cite{bahdanau2014neural} induced over these contextualized embeddings reflect the importance not of individual inputs, but rather of unknown interactions between all input tokens.
This makes it misleading to present heatmaps of these scores over the original token inputs as an explanation for a prediction~\cite{jain-wallace-2019-attention, wiegreffe-pinter-2019-attention, serrano-smith-2019-attention}.

The key missing property here is \emph{faithfulness}~\cite{lipton2016mythos}: An explanation provided by a model is faithful if it reflects the information actually used by said model to come to a disposition. 
In some settings the ability of a model to provide faithful explanations may be paramount.
For example, without faithful explanations, we cannot know whether a model is exploiting sensitive features such as gender~\cite{pruthi2019learning}. 

We propose an approach to neural text classification that provides faithful explanations for predictions by construction. 
Following prior work in this direction~\cite{lei2016rationalizing}, we decompose our model into independent extraction and prediction modules, such that the latter uses only inputs selected by the former. This discrete selection over inputs allows one to use an arbitrarily complex prediction network while still being able to guarantee that it uses only the extracted input features to inform its output. 

The main drawback to this rationalization approach has been the difficulty of training the two components jointly under only instance-level supervision (i.e., without token labels).
This has necessitated training the extraction module via reinforcement learning --- namely REINFORCE~\cite{williams1992simple} --- which exhibits high variance and is particularly sensitive to choice of hyperparameters. Recent work~\cite{bastings-etal-2019-interpretable} has proposed a differentiable mechanism to perform binary token selection, but this relies on the \emph{reparameterization trick}, which similarly complicates training. Methods using the reparameterization trick tend to zero out token embeddings, which may adversely affect training in transformer-based models, especially when one is not fine-tuning lower layers of the model due to resource constraints, as in our experiments.  

To avoid the complexity inherent to training under a remote supervision signal, we introduce \textbf{\methodfullname{}} (\textbf{\methodname{}}), which disconnects the training regimes of the extractor and predictor networks, allowing each to be trained separately. We still assume only instance-level supervision; the trick is to define a method of selecting snippets from inputs --- \emph{rationales} \cite{zaidan2007using} --- that can be used to support prediction. 
Here we propose using arbitrary feature importance scoring techniques to do so.
Notably, these need not satisfy the `faithfulness' criterion. 
\begin{figure*}
\centering
\includegraphics[width=.9\linewidth]{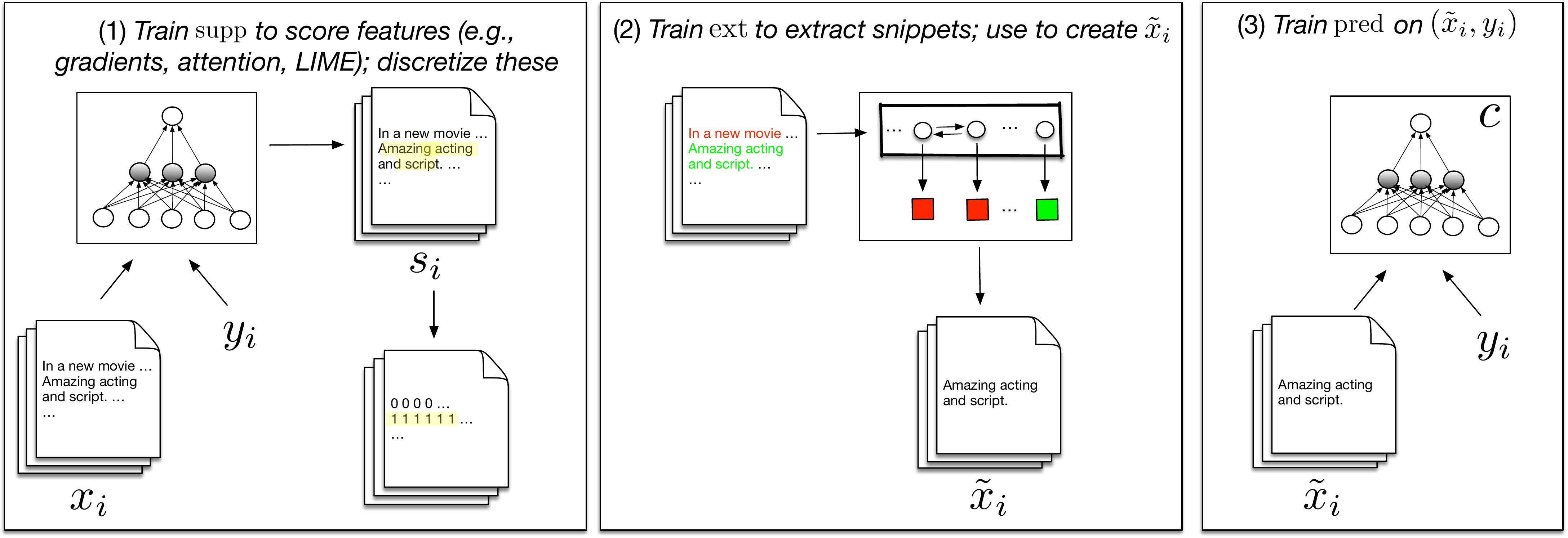} 
\caption{A schematic of the \methodname{} approach. (1) The first model, $\mone$, is trained end-to-end for prediction but used only to `importance score' features. These scores can be derived via any method, e.g., gradients or attention, and are not required to faithfully explain model outputs. Scores are heuristically discretized into binary labels. (2) An extraction module $\rat$ may be a parameterized sequence tagging model trained on the pseudo-targets derived in (1), or heuristics over importance scores directly, creating a new dataset $\langle\tilde{x},y\rangle$ comprising pairs of extracted rationales only. (3) This new dataset is used to train a final classifier, $\mtwo$, which only ever sees rationales.} 
\label{figure:schematic}
\end{figure*}

In this paper we evaluate variants of \methodname{} that use attention ~\cite{bahdanau2014neural} and gradient methods~\cite{li2015visualizing,simonyan2013deep} as illustrative feature scoring mechanisms.
These provide continuous scores for features; we derive discrete rationales from them using simple heuristics. An independent network then uses \emph{only} the extracted rationales to make predictions. 

Disconnecting the training tie between the independent rationale extractor and prediction modules means that \methodname{} is faithful by construction: The snippet that is ultimately used to inform a prediction can be presented as a faithful explanation because this was the only text available to the predictor.
In contrast to prior discrete rationalization methods, \methodname{} greatly simplifies training, and can accommodate any feature importance scoring metric. 
In our experiments, we also find that it yields superior predictive performance. 

In addition to being faithful (and affording strong predictive performance), extracted rationales would ideally be intuitive to humans, i.e., \emph{plausible}. To evaluate this we run a small user study (\autoref{sec:human}) in which humans both evaluate the readability of extracted rationales and attempt to classify instances based on them, effectively serving as a prediction module in the $\methodname{}$ framework. An example illustrating this property is presented in \autoref{fig:example}.

\section{Related Work} 

\paragraph{Types of explainability.}
\newcite{lipton2016mythos, doshi2017towards} and \newcite{rudin2018please} provide overviews on definitions and characterizations of interpretability. \newcite{plkumjorn2019evaluations} classify three possible uses of text explanations: (\emph{i}) revealing model behavior, (\emph{ii}) justifying model predictions, and (\emph{iii}) helping humans investigate uncertain predictions. Attempting to guarantee the faithfulness of a feature selection or explanation generation method is a more challenging question than finding explanations which humans find acceptable~\cite{rudin2018please}. But the benefits of developing such methods is profound: Faithful explanations provide a means to reveal a model's underlying decision-making process.

\paragraph{Issues with current explainability methods in NLP.} A recent line of work in NLP has begun to critically examine the use of certain methods for constructing `heatmaps' over input tokens to explain predictions. In particular, existing feature attribution methods may not provide robust, faithful explanations \cite{feng2018pathologies, jain-wallace-2019-attention,wiegreffe-pinter-2019-attention,serrano-smith-2019-attention,brunner2019identifiability,zhong2019fine,pruthi2019learning}. 

\newcite{wiegreffe-pinter-2019-attention} argue for classifying model interpretability into two groups: faithfulness and plausibility. \newcite{lei2016rationalizing} note that a desirable set of criteria for rationales is that they are sufficient, short, and coherent. \newcite{yu2019rethinking} extend these criteria by additionally arguing for comprehensiveness, which dictates that a rationale should contain all relevant and useful information.

Prior efforts \cite{lei2016rationalizing, yu2019rethinking, bastings-etal-2019-interpretable} have proposed methods that produce faithful explanations via a two-model setup, defining a \emph{generator} network that imposes \emph{hard attention} over inputs and then passes these to a second model for prediction. \citet{yu2019rethinking} extend this by adding a third adversarial model into the framework. These models are trained jointly, which is difficult because hard attention is discrete and necessitates recourse to reinforcement learning, i.e., REINFORCE \cite{williams1992simple}, or the reparameterization trick \cite{bastings-etal-2019-interpretable}. 

\paragraph{Human evaluations.}
\citealt{kim2016examples} states: \say{a method is interpretable if a user can correctly predict the method's result}; they conducted user studies to test this. In a similar plausibility vein, others have proposed testing whether humans \emph{like} rationales \cite{ehsan2018rationalization, ehsan2019automated}. 
We follow these efforts by eliciting human judgments on rationales, although we view plausibility as a secondary aim here.

\section{Faithfulness through Discrete Rationale Selection}
\label{sec:methods}

We now propose \methodname{}, our framework for training explainable neural predictors. 
We begin by describing the two-model, discrete rationale selection approach introduced by~\citet{lei2016rationalizing} (\S\ref{section:lei}), which serves as the starting point for our framework, detailed in \S\ref{section:ours}. 

\subsection{End-to-End Rationale Extraction}
\label{section:lei} 

Consider a standard text classification setup in which we have $n$ input documents $X = \{x_1, ... , x_n\}$, $x_i \in V^{l_i}$, where $l_i$ denotes the number of tokens in document $x_i$, and $V$ the vocabulary, and their assigned labels $y = \{y_1, ..., y_n\}$, $y_i \in \mathcal{Y}$.
\citeauthor{lei2016rationalizing} propose a model comprising a generator ($\gen$) and an encoder ($\enc$).
$\gen$ is tasked with extracting rationales from inputs $x_i$, formalized as a binary mask over tokens sampled from a Bernoulli distribution: $z_i\sim\gen(x_i)\in\{0,1\}^{l_i}$.
$\enc$ makes predictions $\hat{y}=\enc(x_i, z_i)$ on the basis of the unmasked tokens. 

The objective function is defined so that the overall expected loss $\mathcal{L}$ is minimized over both modules:
\begin{equation}
    \label{eq:e2eloss}
    \displaystyle{\minimize_{\theta_{\enc}, \theta_{\gen}} \sum_{i=1}^n \mathds{E}_{z_i \sim {\gen(x_i)}} \mathcal{L}\left(\enc(x_i, z_i),y_i\right)}.
\end{equation}

This objective (\ref{eq:e2eloss}) is difficult to optimize as it requires marginalizing over all possible rationales $z$.
Parameter estimation is therefore performed via an approximation approach that entails drawing samples from $\gen(x)$ and averaging their associated gradients during the learning process.
\newcite{lei2016rationalizing} found that this REINFORCE-style estimation works well for rationale extraction, but may have high variance as a result of the large state space of possible rationales under consideration, which is difficult to efficiently explore.

The loss function $\mathcal{L}$ used by~\citet{lei2016rationalizing} is a squared $\ell_2$ loss between the prediction $\enc(x,z)$ and the reference label $y$, with added regularization terms placed on the binary mask $z$ to encourage rationale conciseness and contiguity.

We modify the conciseness term so that the model is not penalized as long as a predefined desired rationale length $d$ has not been passed:
\begin{equation}
\small
    \Omega(z) = \lambda_1 \underbrace{\max\left(0, \frac{|z|}{L} - d\right)}_{\text{conciseness}} + \lambda_2 \underbrace{\sum_t {\frac{|z_t - z_{t-1}|}{L - 1}}}_{\text{contiguity}}.
    \label{eq:ourgularizer}
\end{equation}

\section{\methodfullname{} (\methodname)}
\label{section:ours}

To avoid recourse to REINFORCE, we introduce \methodname{}, in which we decompose the original prediction task into three sub-components, each with its own independent model.
These are the \textbf{support} model $\mone$, the rationale \textbf{extractor} model $\rat$, and the \textbf{classifier} $\mtwo$.\footnote{This is the most general framing, but in fact $\mone$ and $\rat$ may be combined by effectively defining $\rat$ as an application of heuristics to extract snippets on the basis of scores provided by $\mone$; any means of procuring `importance' scores for the features comprising instances and converting these to extracted snippets to pass to $\mtwo$ will suffice.} 

We train $\mone$ end-to-end to predict $y$, using its outputs only to extract continuous feature importance scores from instances in $X$. These scores are binarized  by $\rat$ either using a parameterized model trained on the output scores, or via direct discretization heuristics. Finally, $\mtwo$ is trained (and tested) \emph{only on text provided by $\rat$}.
Figure \ref{figure:schematic} depicts this proposed framework.

A central advantage of our decomposed setup lies in the arbitrariness of the rationale extraction mechanism.
Any function over $\mone$'s predictions that assigns scores to the input tokens intended to quantify their importance can serve as an input to $\rat$. Note that this means even post-hoc scoring models (applied after the model has completed training) are permissible. 
Examples of such functions include gradient-based methods and LIME~\cite{ribeiro2016should}.

Notably, the importance scoring function need not faithfully identify features that actually informed the predictions from $\mone$.
This means, e.g., that one is free to use token-level attention (over contextualized representations) --- the final rationales provided by FRESH will nonetheless remain faithful with respect to $\mtwo$. 
The importance scores are used only to train $\rat$ heuristically, for example by treating the top $k$ tokens (with respect to importance scores) for a given example as the target rationale. 
The key design decision here is designing such heuristics that map continuous importance scores to discrete rationales.
Any strategy for this will likely involve trading conciseness (shorter rationales) against performance (greater predictive accuracy).

For explainability, we can present users with the snippet(s) that $\mtwo$ used to make a prediction as an explanation (from $\rat$), and we can be certain that the only tokens that contributed to the prediction made by $\mtwo$ are those included in the this text.
In addition to transparency, this framework may afford efficiency gains in settings in which humans are tasked with classifying documents; in this case we can use $\rat$ to present only the (short) relevant snippets. Indeed, we use exactly this approach as one means of evaluation in Section~\ref{sec:human}.

\vspace{-0.3em}
\section{\methodname{} Implementations}
\vspace{-0.3em}
The high-level framework described above requires making several design choices to operationalize; we propose and evaluate a set of such choices in this work, detailed below. Specifically, we must specify a feature importance scoring mechanism for $\mone$ (Section \ref{section:feature-scoring}), and a strategy for inducing discrete targets from these continuous scores (\ref{section:discretize}). In addition, we need to specify a trained or heuristic extractor architecture $\rat$. In this work, all instances of $\mtwo$ exploit BERT-based representations.\footnote{For fair comparison, we have modified all baselines \cite{lei2016rationalizing,bastings-etal-2019-interpretable} to similarly capitalize on BERT-based representations.}

\subsection{Feature Scoring Methods}
\label{section:feature-scoring}

All models considered in this work are based on Bidirectional Encoder Representations from Transformer (BERT) encoders~\cite{devlin-etal-2019-bert} and its variants, namely RoBERTa~\cite{liu2019roberta} and SciBERT~\cite{Beltagy2019SciBERT}; see \autoref{appendix:hyperparams-lei} for more details. For sake of brevity, we simply refer to all of these as BERT from here on. 
We define $\mone$ as a BERT encoder that consumes either a single input (in the case of standard classification) or two inputs (e.g., in the case of question answering tasks) separated by the standard {\tt [SEP]} token. 

While we emphasize that the proposed framework can accommodate arbitrary input feature scoring mechanisms, we consider only a few obvious variants here, leaving additional exploration for future work.
Specifically, we evaluate attention scores~\cite{bahdanau2014neural} and input gradients~\cite{li2015visualizing,simonyan2013deep}. 

Attention scores are taken as the self-attention weights induced from the {\tt [CLS]} token index to all other indices in the penultimate layer of $\mone$; this excludes weights associated with any special tokens added. BERT uses wordpiece tokenization; to compute a score for a token, we sum the self-attention weights assigned to its constituent pieces. BERT is also multi-headed, and so we average scores over heads to derive a final score.

\subsection{Discretizing Soft Scores}
\label{section:discretize} 

A necessary step in our framework consists of mapping from the continuous feature scores provided by $\mone$ to discrete labels, or equivalently, mapping scores to rationales which will either be consumed directly by $\mtwo$ or be used to train a sequence tagging model $\rat$. We consider a few heuristic strategies for performing this mapping. 

\paragraph{Contiguous.}  Select the span of length $k$ that corresponds to the highest total score (over all spans of length $k$). We call these rationales \textbf{contiguous}. 

\paragraph{Top-$k$.} Extract as a rationale the top-$k$ tokens (with respect to importance scores) from a document, irrespective of contiguity (each word is treated independently). We refer to these rationales as \textbf{non-contiguous}. 

\vspace{.5em}
\noindent These strategies may be executed \textbf{per-instance} or \textbf{globally} (across an entire dataset), reflecting the flexibility of \methodname{}.
Empirically, per-instance and global approaches performed about the same; we report results for the simpler, per-instance approaches (additional results in \autoref{app:global}).

\subsection{Extractor model}
\label{section:extractor} 

We experiment with two variants of $\rat$. The first is simply direct use of the importance scores provided by $\mone$ and discretization heuristics over these; this does not require training an explicit $\rat$ model.
We also consider a parameterized extractor model that independently makes token-wise predictions from BERT representations. Using an explicit extraction model allows us to mix in direct supervision on rationales alongside the pseudo-targets derived heuristically from $\mone$. 

Tying the sequential token predictions made by $\rat$ via a Conditional Random Field (CRF) layer ~\cite{lafferty2001conditional} may further improve performance, but we leave this for future work. 

\section{Experimental Setup}

\subsection{Datasets}

\begin{table}
    \centering
    \small
    \begin{tabular}{lrcr}
        \toprule
        {} &  Doc. Len.   & Rationale Len. & $N$ \\
        \midrule
        SST &  17 &  - &  9,613 \\
        AGNews &  30 &  - &  127,600 \\
        Ev. Inf. & 349 &  10\% &  7,193 \\
        Movies &  728 & 31\% &  1,999 \\
        MultiRC &  297 &  18\% &  32,091 \\
        \bottomrule
    \end{tabular}
    \caption{Dataset details, with rationale length ratios included for datasets where they are available.
    }
    \label{tab:dataset-stats}
\end{table}

\begin{table*}[htbp!]
    \centering
    \small
    \begin{tabular}{llrrrrr}
        \toprule
        Saliency & Rationale  &                SST (20\%) &            AGNews (20\%) &             Ev. Inf. (10\%) &            Movies (30\%) &           MultiRC (20\%)\\
        \midrule
        \emph{Full text} & -- &  .90 (.89-.90) &  .94 (.94-.94) &   .73 (.73-.78) &  .95 (.93-.97) &   .68 (.68-.69) \\
        \midrule
        \multirow{2}{*}{Lei \emph{et al.}} & contiguous &  .71 (.49-.83) &  .87 (.85-.89) &   .53 (.45-.56) &  .83 (.80-.92) &   .62 (.62-.64) \\
          & top $k$  &  .74 (.47-.84) &  \textbf{.92 (.90-.92)} &  .47 (.38-.53) &  .87 (.80-.91) &   .64 (.61-.65) \\
        \midrule
          \multirow{2}{*}{Bastings \emph{et al.}} & contiguous & .60 (.58-.62) &  .77 (.18-.78) &  .45 (.40-.49) & --- & .41 (.30-.50) \\
     & top $k$ & .59 (.58-.61) &  .72 (.19-.80) &  .50 (.38-.60) & --- & .44 (.30-.55) \\
        \midrule
        \multirow{2}{*}{Gradient} & contiguous  &  .70 (.69-.72) &  .85 (.84-.85) &  .67 (.62-.68) &  \textbf{.94 (.92-.95)} &  \textbf{.67 (.66-.67)}\\
                & top $k$  &  .68 (.67-.70) &  .86 (.85-.86) &  .62 (.61-.64) &  .93 (.92-.94) &   .66 (.65-.67) \\
        \multirow{2}{*}{{\tt [CLS]} Attn} & contiguous  &  \textbf{.81 (.80-.82)} &  .88 (.88-.89) &  \textbf{.68 (.59-.73)} &  .93 (.90-.94) &  .63 (.60-.62) \\
                & top $k$ &   \textbf{.81 (.80-.82)} &  \textbf{.91 (.90-.91)} & .66 (.64-.70) &  \textbf{.94 (.93-.95)} &  .63 (.62-.64)\\
        
        \bottomrule
    \end{tabular}
    \caption{Model predictive performances across datasets, with rationale length as a percentage of each document in parentheses.
    We report mean Macro F1 scores on test sets, and min/max across random seeds.
    The top row (\emph{Full text}) corresponds to a black-box model that does not provide explanations and uses the entire document; this is upper-bound on performance.
    We bold the best-performing rationalized model(s) for each corpus.}
    \label{table:main-results}
\end{table*}

We use five \textbf{English} text classification datasets spanning a range of domains (see \autoref{tab:dataset-stats}).
\paragraph{Stanford Sentiment Treebank (SST)~\cite{socher2013recursive}.} Sentences labeled with binary sentiment (neutral sentences have been removed).
\paragraph{AgNews \cite{del2005ranking}.} News articles to be categorized topically into \emph{Science}, \emph{Sports}, \emph{Business}, and \emph{World}.
\paragraph{Evidence Inference~\cite{lehman-etal-2019-inferring}.} Biomedical articles describing randomized controlled trials.
The task is to infer the reported relationship between a given intervention and comparator with respect to an outcome, and to identify a snippet within the text that supports this.
The original dataset comprises lengthy full-text articles; we use an abstract-only subset of this data.
\paragraph{Movies~\citep{zaidan2008modeling}.} Movie reviews labeled for sentiment accompanied by rationales on dev and test sets ~\cite{deyoung2019eraser}.
\paragraph{MultiRC~\citep{KCRUR18-multirc}.}
Passages and questions associated with multiple correct answers.
Following~\citet{deyoung2019eraser}, we convert this to a binary classification task where the aim is to categorize answers as \emph{True} or \emph{False} based on a supporting rationale.

\subsection{Model and Training Details}
\label{ssec:k}
For datasets where human rationale annotations are available, we set $k$ to the average human rationale annotation length, rounded to the nearest ten percent.
For the rest, we set $k=20\%$. 

For generality, all models considered may consume both \emph{queries} and texts, as is required for MultiRC and Evidence Inference.
Rationales can be extracted from only from the text; this typically dominates the query in length, and is more informative in general.
Further implementation details (including hyperparameters) are provided in \autoref{appendix:details}.

\paragraph{Hyperparameter sensitivity and variance.} To achieve conciseness and contiguity,~\citet{lei2016rationalizing} impose a regularizer on the encoder that comprises two terms (\autoref{eq:ourgularizer}) with associated hyperparameters ($\lambda_1$, $\lambda_2$). In practice, we have found that one needs to perform somewhat extensive hyperparameter search for this model to realize good performance. This is inefficient both in the sense of being time-consuming, and in terms of energy~\cite{strubell-etal-2019-energy}.

By contrast, \methodname{} requires specifying and training independent module components, which incurs some energy cost. But there are no additional hyperparameters, and so \methodname{} does not require extensive hyperparameter search, which is typically the most energy-intensive aspect of model training. We quantify this advantage by reporting the variances over different hyperparameters we observed for~\cite{lei2016rationalizing} and the compute time this required to conduct this search in \autoref{appendix:hyperparams-lei}.

In addition to being sensitive to hyperparameters, a drawback of REINFORCE-style training is that it can exhibit high variance within a given hyperparameter setting.
To demonstrate this, we report the variance in performance of our proposed approach and of \citet{lei2016rationalizing} as observed over five different random seeds. 

We also find that both \citet{lei2016rationalizing} and \citet{bastings-etal-2019-interpretable} tend to degenerate and predict either complete or empty text as rationale. To make results comparable to \methodname{}, at inference time, we restrict the rationale to specified desired length $k$ before passing it to the corresponding classifier.

\section{Quantitative Evaluation}
\label{section:results}

\begin{figure*}  
    \centering
    \includegraphics[width=.825\linewidth]{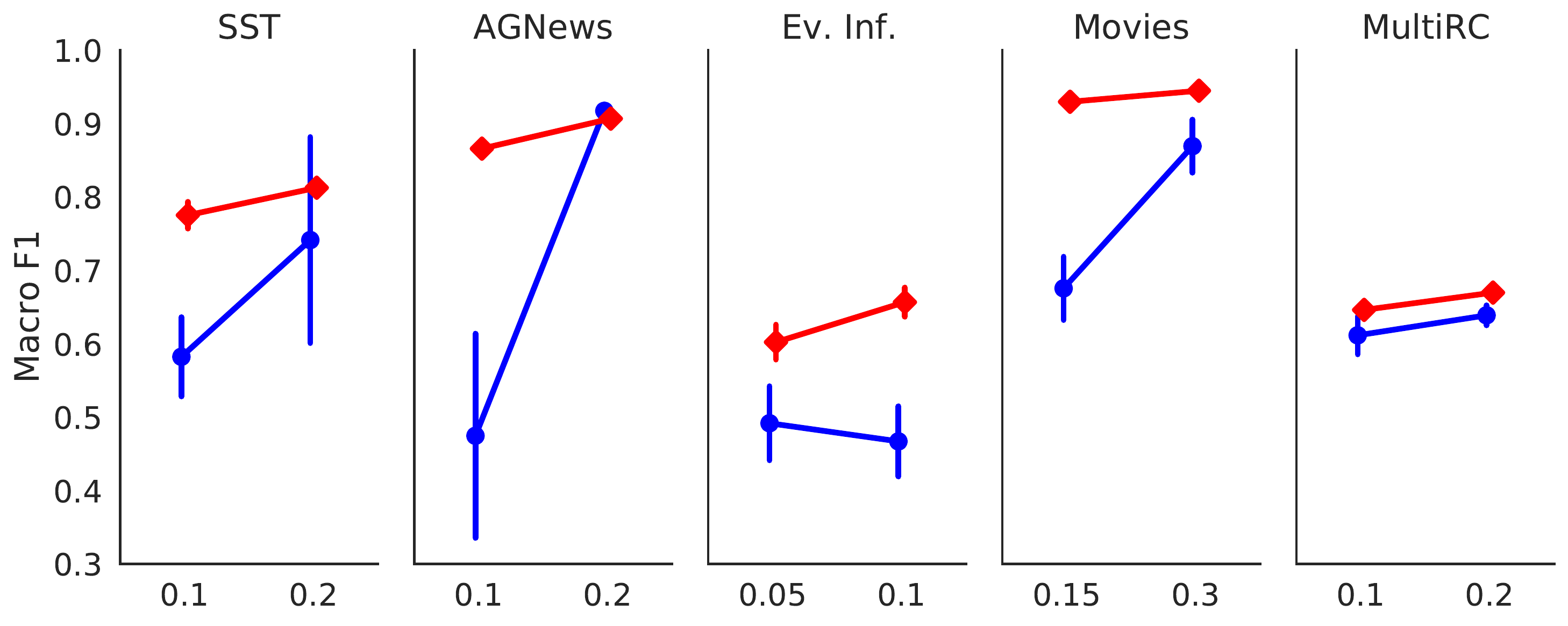}
    \caption{Results for Lei \emph{et al.} (\tikz\draw[blue,fill=blue] (0,0) circle (.5ex);) and \methodname{} ($\MyDiamond[draw=red,fill=red]$) evaluated across five datasets at two different desired rationale lengths (as \% of document length). Vertical bars depict standard deviations observed over five random seeds.}
    \label{fig:cutpoints}
\end{figure*}

We first evaluate the performance achieved on datasets by the $\mtwo$ models trained on different $\rat$-extracted rationales, compared to each other and to~\newcite{lei2016rationalizing}'s end-to-end rationale extraction framework. 
As an additional baseline, we also evaluate a variant of the differentiable binary variable model proposed in \citealt{bastings-etal-2019-interpretable}. This baseline do not require any hyperparameter search. 

In general, we would expect predictive performance to positively correlate with rationale length, and so we evaluate predictive performance (accuracy or F1-score) across methods using a fixed rationale length for each dataset.

We report results in terms of predictive performance for all model variants in~\autoref{table:main-results}.
Here we use the entire train sets for the respective datasets, and fix the rationale length as described in \S\ref{ssec:k} to ensure fair comparison across methods.
We observe that despite its simplicity, \methodname{} performs nearly as well as \textit{Full text} while using only 10-30\% of the original input text, thereby providing transparency. \methodname{} achieves better average performance than \citeauthor{lei2016rationalizing}'s end-to-end method, with the exception of AGNews, in which case the models are comparable. It also consistently fares better than \citeauthor{bastings-etal-2019-interpretable}'s system.

Of the two feature scoring functions considered, {\tt [CLS]} self-attention scores tend to yield better results, save for on the MultiRC and Movies datasets, on which gradients fare better.
With respect to discretizing feature scores, the simple top-$k$ strategy seems to perform a bit better than the contiguous heuristic, in what we expect to be traded off against a greater coherence of the contiguous rationales. 

As seen in~\autoref{table:main-results}, \methodname{} exhibits lower variance across runs, and does not require hyperparameter search (further analysis in \autoref{appendix:hyperparams-lei}). 

\paragraph{Varying rationale length.}\autoref{fig:cutpoints} plots F1 scores across datasets and associated standard deviations achieved by the best rationale variant of \citet{lei2016rationalizing} and \methodname{} at two different target rationale lengths. These results demonstrate the effectiveness of \methodname{} even in constrained settings. Note, we had to re-perform hyperparameter search for a different rationale length in case of~\cite{lei2016rationalizing} model.

\paragraph{Incorporating human rationale supervision.} In some settings it may be feasible to elicit direct supervision on rationales, at least for a subset of training examples. Prior work has exploited such signal during training \cite{zhang2016rationale,Strout2019DoHR,small2011constrained}. One of the potential advantages of explicitly training the extraction model $\rat$ with pseudo-labels for tokens (derived from heuristics over importance scores) is the ability to mix in direct supervision on rationales alongside these derived targets.

\begin{figure}
    \centering
    \includegraphics[width=0.85\columnwidth]{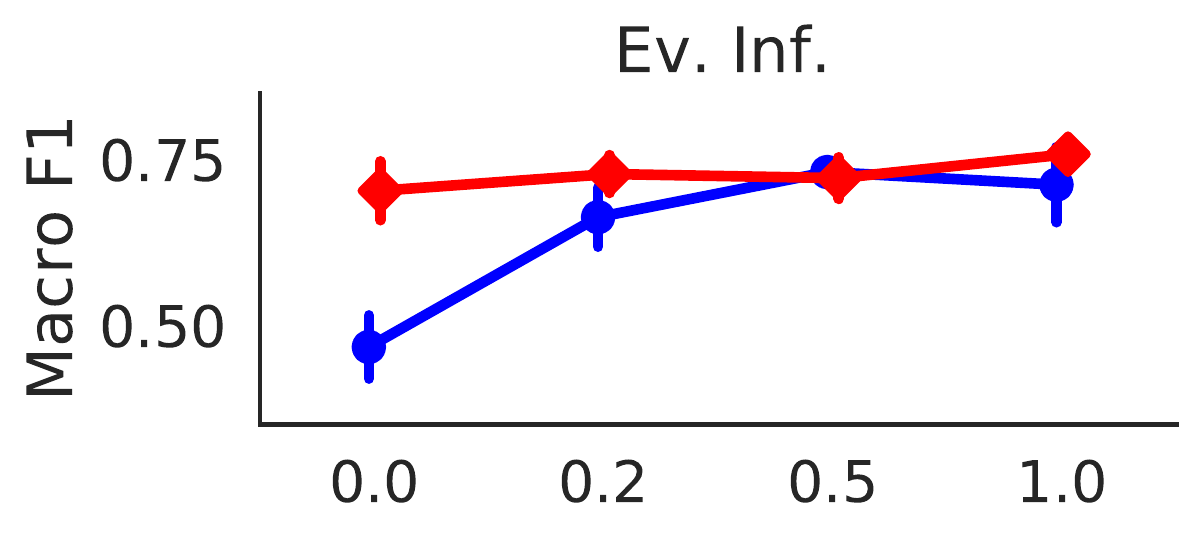}
    \caption{Results on Evidence Inference for  Lei \emph{et al.} (\tikz\draw[blue,fill=blue] (0,0) circle (.5ex);) and \methodname{} ($\MyDiamond[draw=red,fill=red]$) given varying amounts of explicit rationale supervision.}
    \label{fig:rationale-supervision}
\end{figure}

\begin{figure}
    \centering
    \includegraphics[width=0.85\columnwidth]{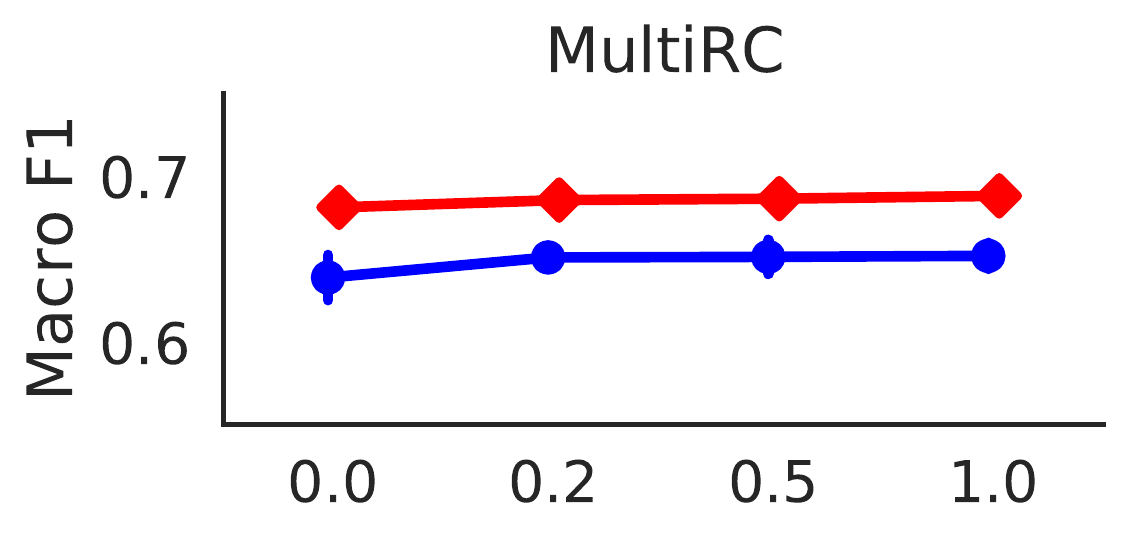}
    \caption{Results on MultiRC for Lei \emph{et al.} (\tikz\draw[blue,fill=blue] (0,0) circle (.5ex);) and \methodname{} ($\MyDiamond[draw=red,fill=red]$) given varying amounts of explicit rationale supervision.}
    \label{fig:rationale-supervision-2}
\end{figure}

We evaluate whether direct rationale supervision improves performance on two datasets for which we have human rationale annotations (Evidence Inference and MultiRC). In both cases we provide models with varying amounts of rationale-level supervision (0\%, 20\%, 50\% and 100\%), and again compare the best variants of \citet{lei2016rationalizing} and our model. For the former, we introduce an additional binary cross entropy term into the objective for that explicitly penalizes the extractor for disagreeing with human token labels.

Explicitly training a sequence tagging model as $\rat$ over heuristic targets from $\mone$ did not improve results in our experiments.
However, as shown in \autoref{fig:rationale-supervision} and \autoref{fig:rationale-supervision-2}, mixing in rationale-level supervision when training $\rat$ \emph{did} improve performance on the Evidence Inference dataset by a small amount, although not for MultiRC. This suggests that explicit rationale supervision may at least sometimes improve performance, and this is not possible without a parameterized $\rat$ model.

In~\citet{lei2016rationalizing}'s framework, direct supervision provides considerable performance improvement in the case of Evidence Inference (although still suffering from variance effects), and did not affect performance on MultiRC.

\section{Human Analysis}\label{sec:human}

We have proposed \methodname{} as an architecture which, in addition to exceeding performance of previous training regimes, provides a guarantee for extracting rationales which are \textit{faithful}.
However, as noted in the introduction, another desirable trait of rationales is that they are judged as \textit{good} by humans.
To assess the plausibility of the resulting rationales~\cite{herman2017promise, wiegreffe-pinter-2019-attention}, we design a human user study.\footnote{We received approval for this study from Northeastern University's Institutional Review Board (IRB).} We evaluate the following attributes of plausibility:

\paragraph{Sufficiency.} Can a human predict the correct label given only the rationale?
This condition aligns with \citealt{kim2016examples}, with \citealt{lei2016rationalizing}, and with the confidence and adequate justification criteria of \citealt{ehsan2019automated}.
In our experiment, we simply substitute a human user for $\mtwo$ and evaluate performance. 

\paragraph{Readability and understandability.} We test the user's preference for a certain style of rationale beyond their ability to predict the correct label.
Our hypothesis is that humans will prefer contiguous to non-contiguous rationales. 
This condition aligns with coherency~\cite{lei2016rationalizing}, human-likeness and understandability~\cite{ehsan2019automated}.

\subsection{Experiments}

We compare extracted rationales on two tasks, Movies and MultiRC, both of which include reference human rationales~\cite{deyoung2019eraser}. We did not choose evidence inference for this set of experiments since the task requires expert knowledge.
Recall that the rationalization task for the Movies dataset involves selecting those words or phrases associated with positive or negative sentiment.
For MultiRC, the rationale must contain sufficient context to allow the user to discern whether the provided answer to the question is true, based on the information in the passage.

We extract rationales, both contiguous and non-contiguous, from 100 randomly-selected test set instances for the following methods: (1) human (reference label) rationales, (2) randomly selected rationales of length $k$, (3) rationales from the best \citealt{lei2016rationalizing} models, and (4) rationales from the best \methodname{} models. 

We present each extracted rationale to three annotators.\footnote{We use Amazon Mechanical Turk for the annotation task, and compensate Turkers at a rate of \$0.24 per HIT. Pay rate is calculated based on the median HIT completion time in a preliminary experiment (2 minutes) and an hourly wage of \$7.20. We require annotators to be within the U.S., but we do not explicitly test for English language proficiency.}
We ask them to perform the following tasks: 
\begin{enumerate}[leftmargin=*]
\itemsep0em 
    \item Classify examples as either \emph{Positive} or \emph{Negative} (Movies), or as \emph{True} or \emph{False} (MultiRC);
    \item Rate their confidence on a 4-point Likert scale from \emph{not confident} (1) to \emph{very confident} (4);
    \item Rate how easy the text is to read and understand on a 5-point Likert scale from \emph{very difficult} (1) to \emph{very easy} (5).
\end{enumerate}

\noindent The first two tasks are designed to evaluate sufficiency, and the third readability and understandability. We provide images of the user interface in \autoref{section:amt}.

We validate the user interface design with gold-label human rationales.
As expected, when using these rationales Turkers are able to perform the labelling task with high accuracy, and they do so with high confidence and readability (first rows of Tables~\ref{tab:human_1} and~\ref{tab:human_2}).
On average, annotators exhibit over 84\% and 89\% inter-annotator agreement on Movies and MultiRC, respectively.\footnote{We assign the majority predicted document label and averaged Likert value for confidence and readability across the 3 annotators for each instance. We report human Accuracy as a measure of how well our annotators have done at predicting the correct document label from only the extracted rationale. All metrics are averaged over the 100 test documents.}

\subsection{Results}

\begin{table}
\small
  \centering
\begin{tabular}{lllll}
\toprule
Rationale  &  Human & Confidence & Readability \\
Source & Acc. & (1--4) & (1--5) \\
\midrule
Human & .99  & 3.44 \textpm{}0.53 & 3.82 \textpm{}0.56\\ \hline
\textbf{Random} & & & \\
Contiguous & .84  & 3.18 \textpm{}0.55 & 3.80 \textpm{}0.57\\
Non-Contiguous & .65  & 2.09 \textpm{}0.51 & 2.07 \textpm{}0.69\\ \hline
\textbf{\citealt{lei2016rationalizing}} & & & \\
 Contiguous  & .88  & 3.39 \textpm{}0.48 & 4.17 \textpm{}0.59\\
 Non-Contiguous  & .84  & 2.97 \textpm{}0.72 & 2.90 \textpm{}0.88 \\ \hline 
\textbf{\methodname{} Best} & & & \\
Contiguous & .92  & 3.31 \textpm{}0.48 & 3.88 \textpm{}0.57\\
 Non-Contiguous & .87  & 3.23 \textpm{}0.47 & 3.63 \textpm{}0.59\\
\bottomrule
\end{tabular}
\caption{Human evaluation results for Movies.}
\label{tab:human_1}
\end{table}

\begin{table}
\small
  \centering
\begin{tabular}{llll}
\toprule
Rationale & Human & Confidence & Readability \\
Source &Acc. & (1--4) & (1--5) \\
\midrule
Human&	.87&	3.50 \textpm{}0.47	 &4.16 \textpm{}0.54\\ \hline
\textbf{Random} & & & \\
Contiguous&	.65&	2.85 \textpm{}0.76 &	3.49 \textpm{}0.74\\
Non-Contiguous&	.58&	2.56 \textpm{}0.68 &	2.39 \textpm{}0.73\\ \hline
\textbf{\citealt{lei2016rationalizing}} & & & \\
Contiguous&	.57 & 2.90 \textpm0.58 & 3.63 \textpm 0.71\\
Non-Contiguous&	.66 & 2.45 \textpm0.67 & 2.19 \textpm 0.75\\ \hline 
\textbf{\methodname{} Best} & & & \\
Contiguous&	.69 & 2.78 \textpm0.67 & 3.68 \textpm 0.6\\
Non-Contiguous&	.65 & 2.60 \textpm0.68 & 2.50 \textpm 0.83\\
\bottomrule
\end{tabular}
\caption{Human evaluation results for MultiRC.}
\label{tab:human_2}
\end{table}

We report results in Tables~\ref{tab:human_1} and~\ref{tab:human_2}.
We observe that humans perform comparably to the trained model (\autoref{table:main-results}) at predicting document labels given only the model-extracted rationales. 
Humans perform at least as well using our extracted rationales as they do with other methods.
They also exhibit a strong preference for contiguous rationales, supporting our hypothesis.  
Lastly, we observe that confidence and readability are high. Thus while our primary goal is to provide faithful rationales, these results suggest that those provided by \methodname{} are also reasonably plausible.
This shows that faithfulness and plausibility are not mutually exclusive, but also not necessarily correlative. 

\section{Conclusions}

We have proposed \methodfullname{} (\methodname{}), a simple, flexible, and effective method to learn explainable neural models for NLP. Our method can be used with any feature importance metric, is very simple to implement and train, and empirically often outperforms more complex rationalized models.

\methodname{} performs discrete rationale selection and ensures the faithfulness of provided explanations --- regardless of the complexity of the individual components --- by using independent extraction and prediction modules.
This allows for contextualized models such as transformers to be used, without sacrificing explainability (at least at the level of rationales).
Further, we accomplish this without recourse to explicit rationale-level supervision such as REINFORCE or the reparameterization trick; this greatly simplifies training.

We showed empirically that \methodname{} outperforms existing models, recovering most of the performance of the original `black-box' model.
Additionally, we found \methodname{} rationales to be at least as plausible to human users as comparable end-to-end methods.

We acknowledge some important limitations of this work. Here we have considered explainability as an instance-specific procedure. The final explanation provided by the model is limited to the tokens provided by the extraction method. Our framework does not currently support further pruning (or expanding) this token set once the rationale has been selected. 

In addition, while we do have a guarantee under our model about which part of the document was used to inform a given classification, this approach cannot readily say why this specific rationale was selected in the first place. Nor do we clearly understand how the $\mtwo$ uses extracted rationale to perform its classification. We view these as interesting directions for future work. 

\ifaclfinal
    \section*{Acknowledgments}
    We thank colleagues at Georgia Tech and Northeastern for their feedback, including Sandeep Soni, Diyi Yang, Ian Stewart, and Jiaao Chen. We also thank the anonymous reviewers for many useful comments and suggestions.
    
    This work was supported in part by the Army Research Office (W911NF1810328), and by the National Science Foundation (CAREER award 1750978). YP is a Bloomberg Data Science PhD Fellow.
\fi

\bibliography{acl2019,anthology}
\bibliographystyle{acl_natbib}

\appendix

\section{Model Details and Hyperparameters}
\label{appendix:details}

For each model below, we use BERT-base-uncased (for SST, AgNews), Roberta-base (for Multirc), and SciBERT (scivocab-uncased) (for Evidence Inference) embeddings (from {\tt huggingface} library \cite{Wolf2019HuggingFacesTS} as they appear in AllenNLP library \cite{gardner-etal-2018-allennlp}) as corresponding pretrained transformer model. 

Tokenization was performed using tokenizer associated with each pretrained transformer models. Only the top two layers of each model were finetuned. For documents greater than 512 in length, we used staggered position embeddings (for example, if an example of length 1024, the position embeddings used are 1,1,2,2,3,3,...).

\paragraph{Lei \emph{et al.} and Bastings \emph{et al.} Models}

We use transformer model to generate token embeddings (max-pooling embeddings from wordpieces) in the generator, placing a dense classification layer on top to return a binary decision. The encoder model also uses the transformer to encode selected tokens and the start token embedding was used to perform final classification. 

For the movies dataset we used a slightly different model to get around the $O(n^2)$ memory bottleneck. Specifically, we first encode 512 token subsequences with the transformer and then run these through a 128-d BiLSTM on top of transformer embeddings. These embeddings are then used to make token level decisions for generator model. In the encoder model, they are collapsed using additive attention module \cite{bahdanau2014neural} into a single vector prior to the final classification.

We used cross-entropy loss to train the encoder, and the optimization was performed using the Adam Optimizer with a learning rate of 2e-5. For regularization, we used 0.2 dropout after transformer embedding layer and placed an 0.001 $\ell_2$ loss over all weights of our network and a grad norm of 5.0. Models were trained for 20 epochs and we kept the best parameters on the basis of macro-F1 score on dev sets.

Hyperparameter search for \citet{lei2016rationalizing} models was performed over $\lambda_1$ and $\lambda_2$ parameters, with $\lambda_1$ uniformly selected over log scale in range [1e-2, 1e-0] and $\lambda_2$ selected from [0.0, 0.5, 1.0, 2.0]. We performed the hyperparameter search 20 times and selected the best of these on the basis of F1 score on dev sets.

\citet{bastings-etal-2019-interpretable} do not require hyperparameter search since it uses a Lagrangian relaxation based optimisation for its regularizers. We use the same initial hyperparameter settings used by the authors in their codebase. We use the Hard Kumaraswamy distribution as provided by the authors here \url{https://github.com/bastings/interpretable_predictions}.

\paragraph{FRESH}

For all three components of the FRESH model, we used the same transformer-based models as mentioned previously to encode tokens. Classification was performed using start token embeddings. Optimisation was performed using Adam optimizer~\cite{kingma2014adam} with a learning rate of 2e-5. We insert a dropout layer following the BERT embedding layer for regularisation, and impose an 0.001 $\ell_2$ loss over all weights of our network. We also enforce a grad norm of 5.0. The model was trained for 20 epochs, and we again kept the best models with respect to macro F1 scores on the dev sets.

For the movies dataset, we use similar modifications as discussed above.

\section{Hyperparameter sensitivity analysis} 
\label{appendix:hyperparams-lei}

\begin{figure*}
    \centering
    \includegraphics[width=\linewidth]{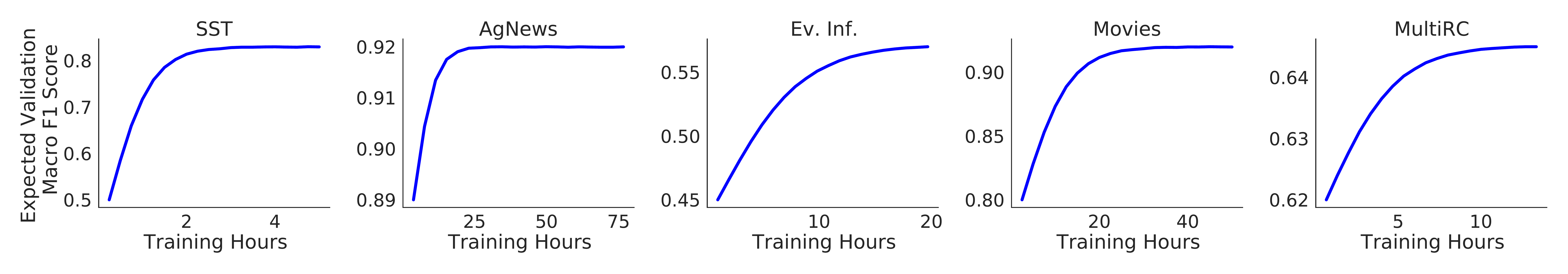}
    \caption{Plot of training duration for a single model vs Expected validation macro F1 scores as defined by \cite{dodge-etal-2019-show}}
    \label{fig:hyperparam-time}
\end{figure*}
\begin{figure}
    \centering
    \includegraphics[width=\columnwidth]{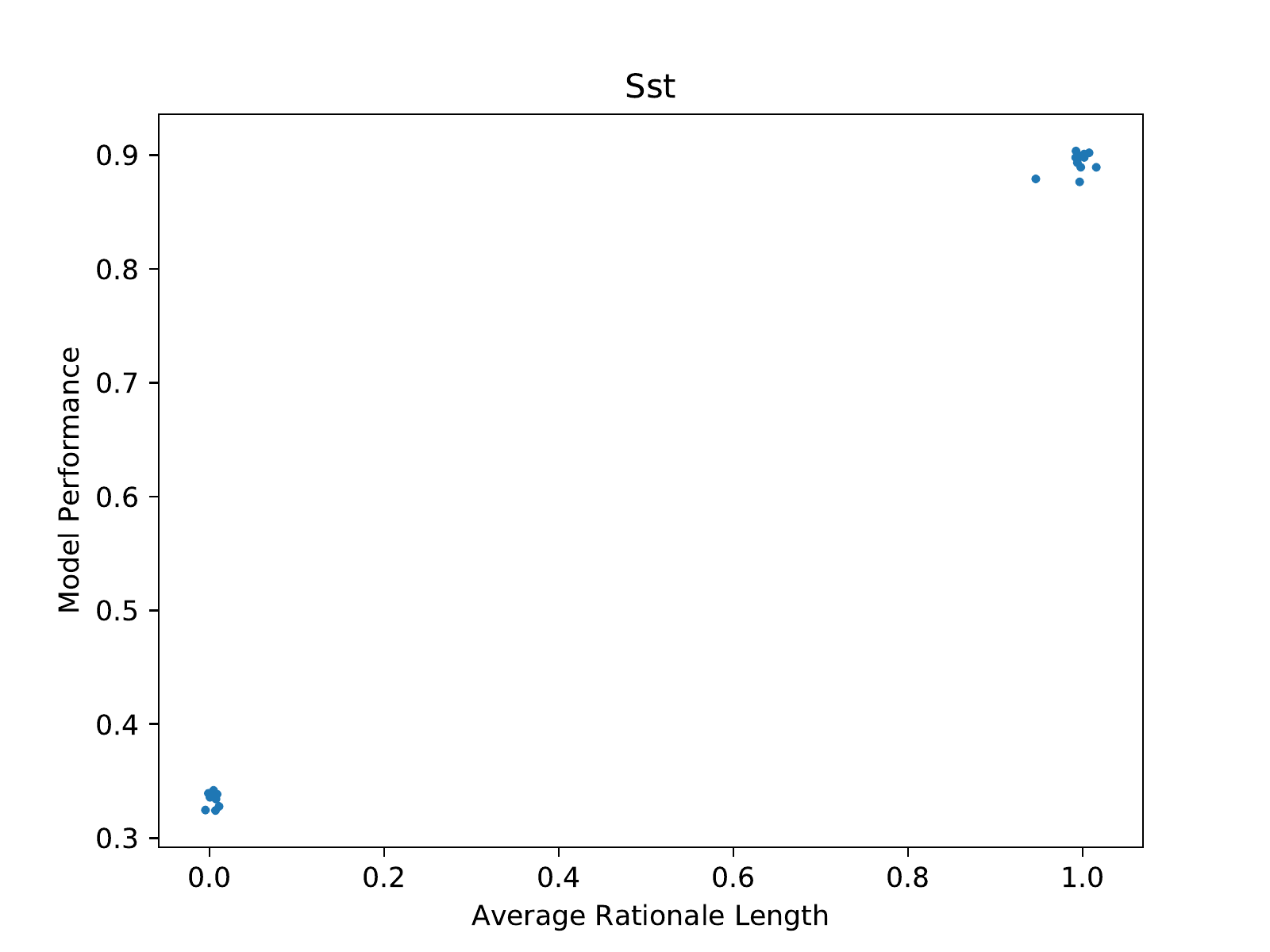}
    \caption{Plot of average rationale length vs model performance in terms of F1 score for 20 hyperparameter searches for~\cite{lei2016rationalizing}}.
    \label{fig:sst}
\end{figure}
\begin{figure}
    \centering
    \includegraphics[width=\columnwidth]{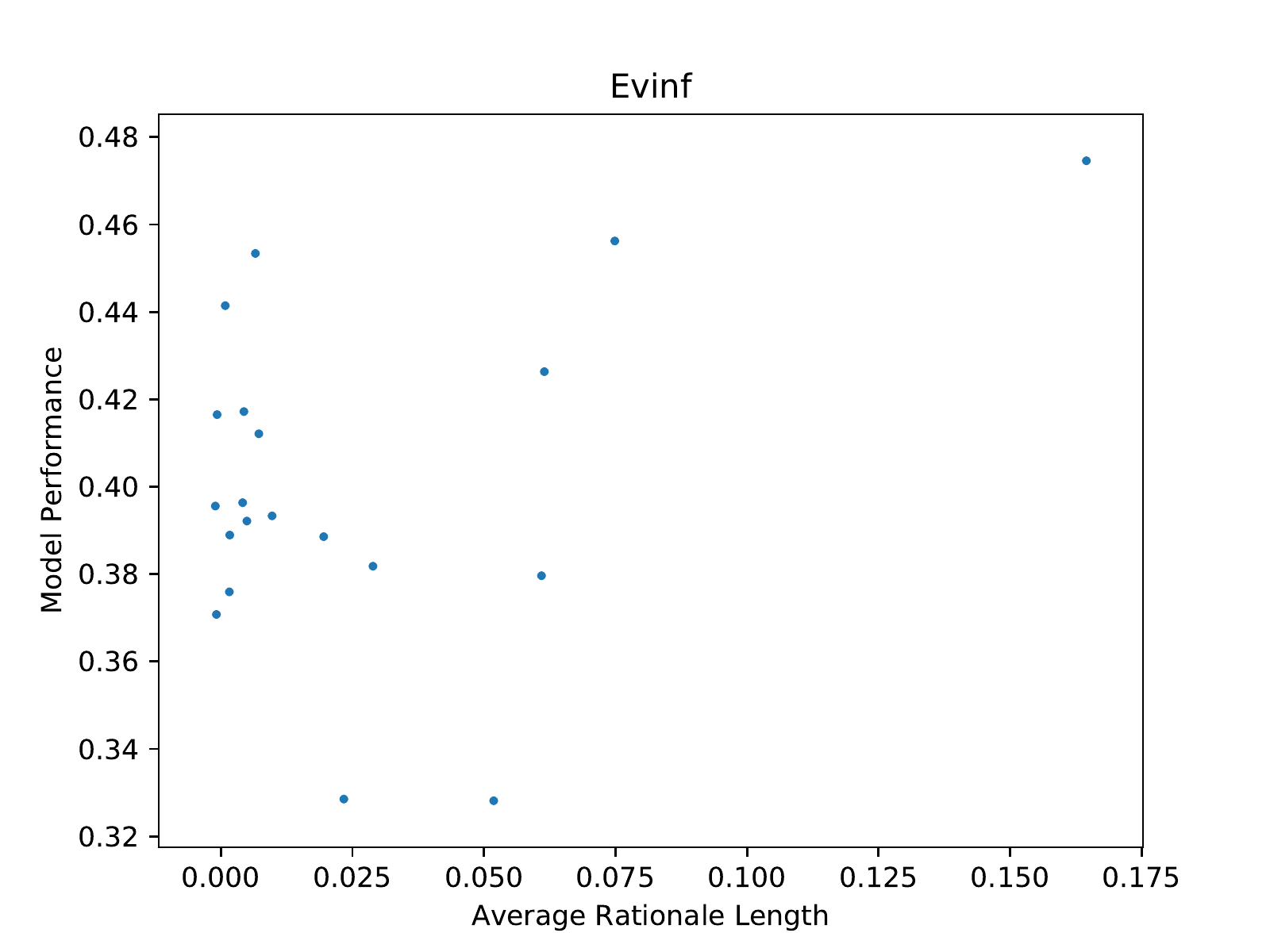}
    \caption{Plot of average rationale length vs model performance in terms of F1 score for 20 hyperparameter searches for~\cite{lei2016rationalizing}}
    \label{fig:evinf}
\end{figure}
\begin{figure}
    \centering
    \includegraphics[width=\columnwidth]{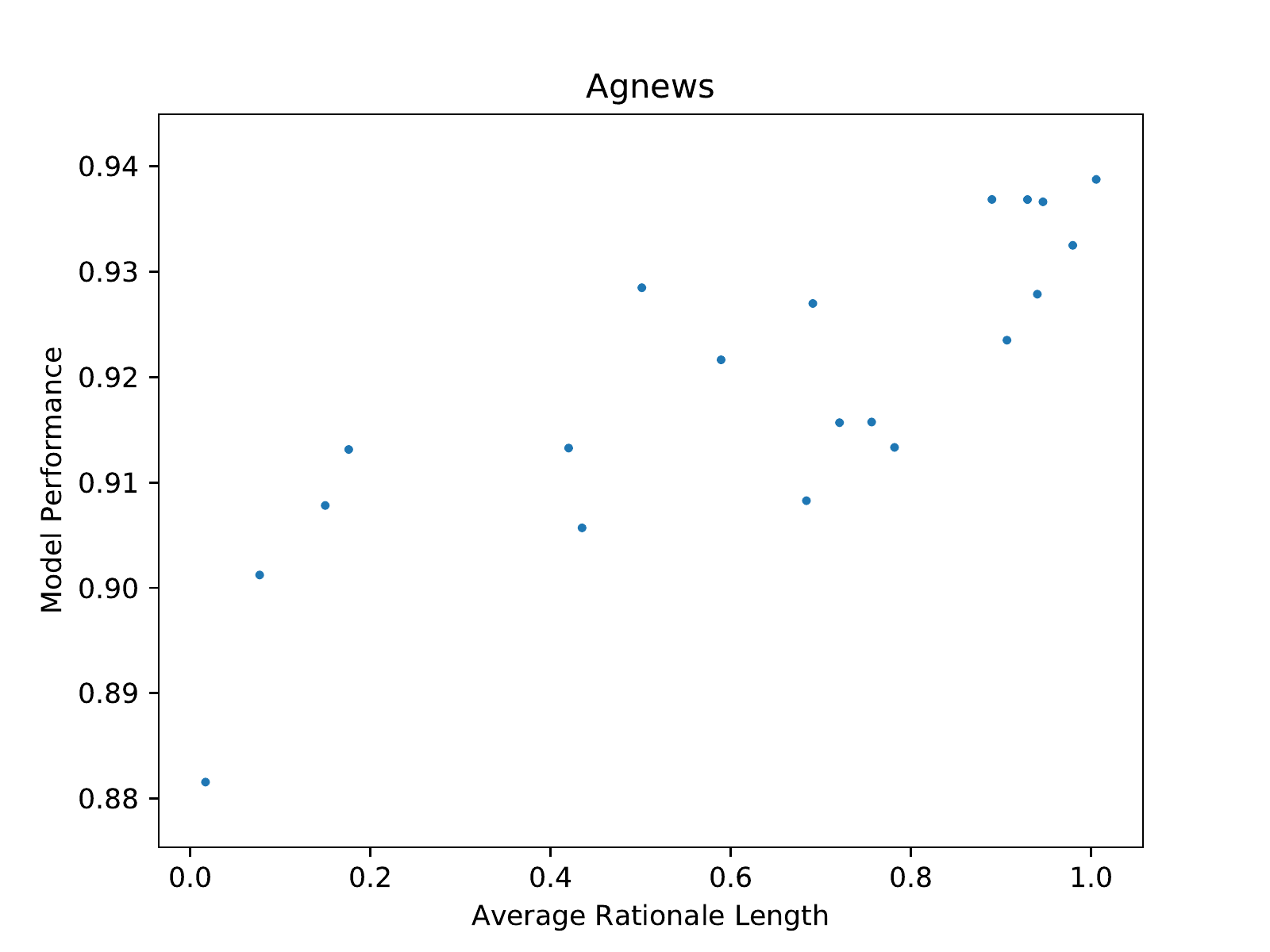}
    \caption{Plot of average rationale length vs model performance in terms of F1 score for 20 hyperparameter searches for~\cite{lei2016rationalizing}}
    \label{fig:agnews}
\end{figure}

In Figures~\ref{fig:sst}, \ref{fig:evinf} and~\ref{fig:agnews}, we report the model accuracy for various hyperparameter searches on three of our datasets.
Note that in many cases, the search does not converge to the desired length (it either selects the entire document or completely degenerates, selecting no tokens).

\section{Amazon Mechanical Turk Layouts}
\label{section:amt}

\begin{figure*}
\includegraphics[width=\linewidth]{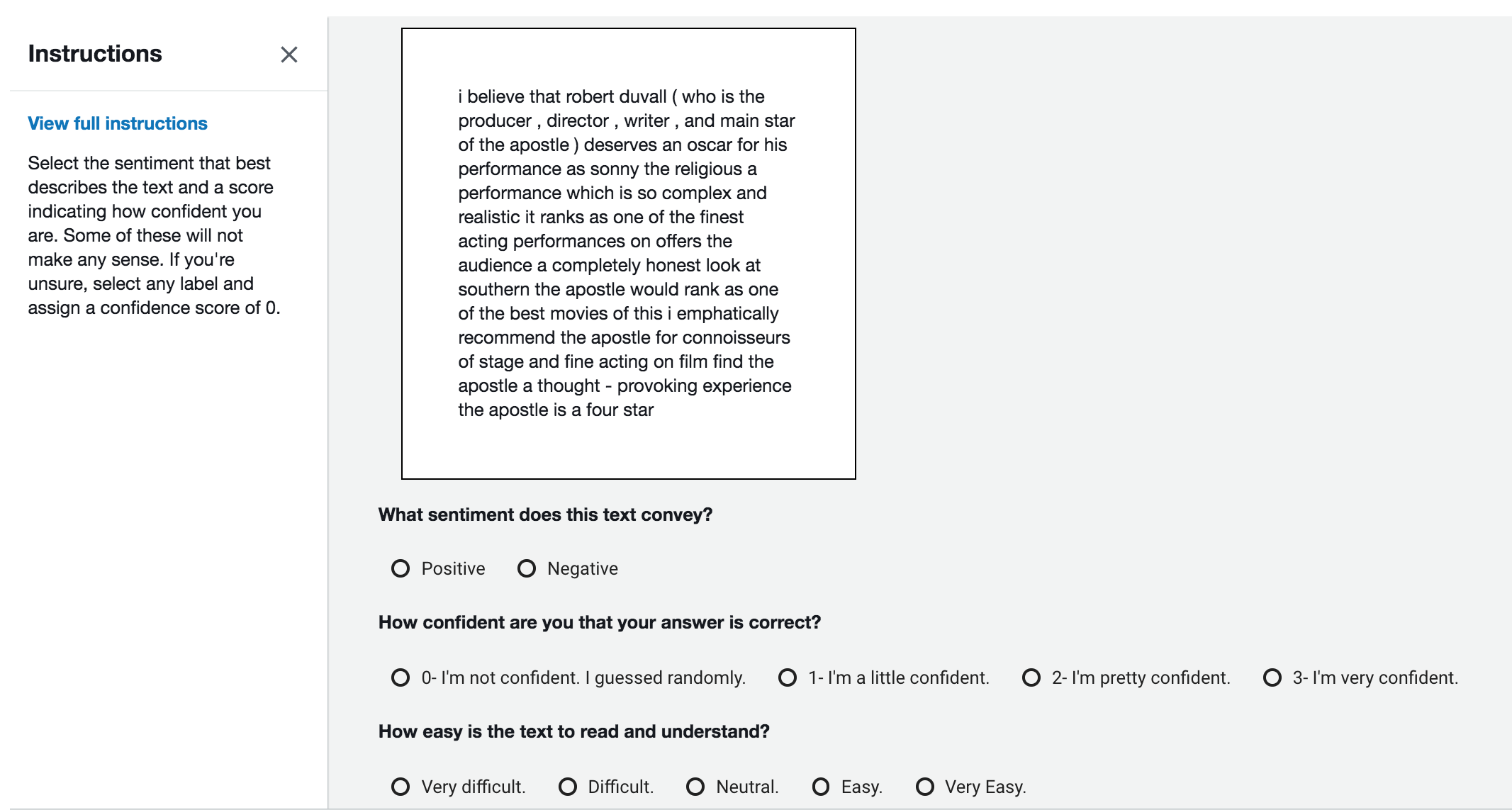} 
\caption{Amazon Mechanical Turk layout for Movies tasks.}
\label{figure:movies_amt}
\end{figure*}

\begin{figure*}
\includegraphics[width=\linewidth]{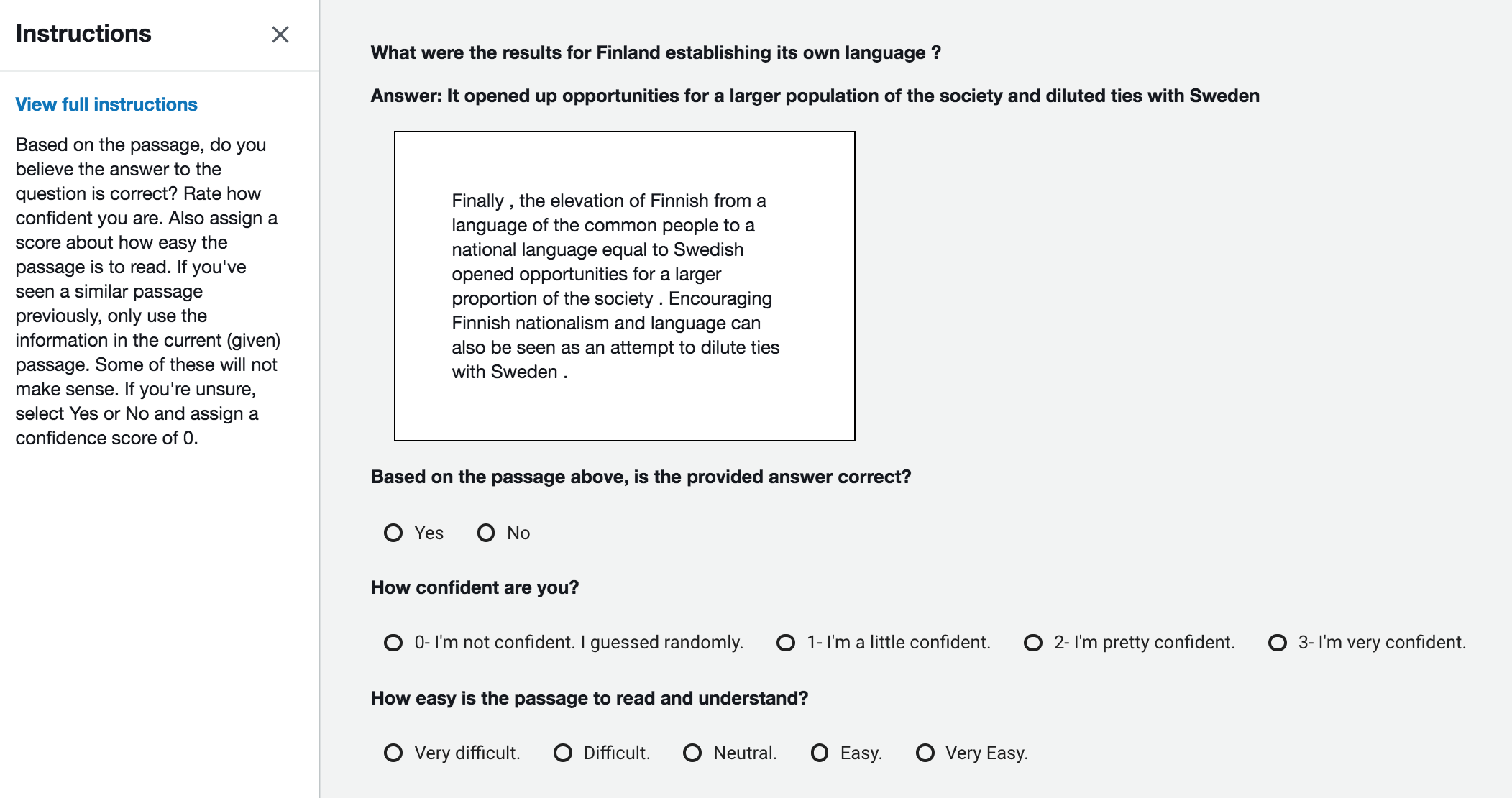} 
\caption{Amazon Mechanical Turk layout for MultiRC tasks.}
\label{figure:multirc_amt}
\end{figure*}

See Figures \ref{figure:movies_amt} and \ref{figure:multirc_amt} for screenshots of the interfaces shown to annotators. 

\section{Additional Dataset Details}

See \autoref{tab:dataset-stats1}.

\begin{table*}
\small
    \centering
\begin{tabular}{llllll}
 &     N &  Doc Length & Query Length & Rationale Length &          Label Distribution \\
\midrule 
\textbf{Evidence Inference} & \multicolumn{5}{l}{} \\ \midrule
{train} &  5,789 &  363 / 1010 &  14 / 66 &  0.10 / 0.54 &  0.39 / 0.33 / 0.28 \\
{dev  } &  684 &  369 / 602 &  14 / 108 &  0.11 / 0.35 &  0.40 / 0.35 / 0.25 \\
{test } &  720 &  362 / 617 &  16 / 100 &  0.10 / 0.34 &  0.39 / 0.35 / 0.26 \\

\midrule 
\textbf{MultiRC} & \multicolumn{5}{l}{} \\\midrule
train &  24,029 &  305 / 618 &  18 / 92 &  0.17 / 0.73 &  0.56 / 0.44 \\
dev   &  3,214 &  305 / 562 &  18 / 83 &  0.19 / 0.76 &  0.55 / 0.45 \\
test  &  4,848 &  290 / 490 &  18 / 80 &  0.18 / 0.56 &  0.57 / 0.43 \\

\midrule 
\textbf{Movies} & \multicolumn{5}{l}{} \\\midrule
train &  1,600 &  773 / 2,809 &  7 / 7 &  0.09 / 0.5 &  0.5 / 0.5 \\
dev   &  200 &  761 / 1,880 &  7 / 7 &  0.07 / 0.26 &  0.5 / 0.5 \\
test  &  199 &  795 / 2,122 &  7 / 7 &  0.31 / 0.91 &  0.5 / 0.5 \\
\midrule 
\textbf{SST} & \multicolumn{5}{l}{} \\\midrule
{Train} & 6,920 & 17 / 48 & - &  - &  0.52 / 0.48 \\
{Dev  } & 872 & 17 / 44 & - &  - &  0.51 / 0.49 \\
{Test } & 1,821 & 17 / 52 & - &  - &  0.50 / 0.50 \\
\midrule 
\textbf{AgNews} & \multicolumn{5}{l}{} \\\midrule
{Train} & 102,000 & 31 / 173 & - &  - &  0.25 / 0.25 / 0.25 / 0.25 \\
{Dev  } & 18,000 & 31 / 168 & - &  - &  0.25 / 0.25 / 0.25 / 0.25 \\
{Test } & 7,600 & 30 / 129 & - &  - &  0.25 / 0.25 / 0.25 / 0.25 \\

\bottomrule
\end{tabular}
    \caption{Dataset statistics. For document, query, and rationale lengths we provide mean and maximum values  (formulated as mean/max), where available. We do not have human rationale annotations for SST and AgNews, hence we do not report query and rationale lengths for these.}
    \label{tab:dataset-stats1}
\end{table*}

\begin{table*}
    \centering
    \small
\begin{tabular}{lllrrr}
\toprule
        &         &       &   $\Delta$ &    t-statistic &    p-value \\
dataset & saliency & rationale &         &          &         \\
\midrule
SST & Gradient & contiguous & -0.0097 &  -1.8483 &  0.1383 \\
        &         & Non contiguous &  0.0120 &   1.9411 &  0.1242 \\
        & \texttt{[CLS]} Attention & contiguous & -0.0133 &  -3.1281 &  0.0352 \\
        &         & Non contiguous & -0.0025 &  -0.5036 &  0.6410 \\
AgNews & Gradient & contiguous & -0.0433 & -26.8053 &  0.0000 \\
        &         & Non contiguous & -0.0014 &  -0.7530 &  0.4934 \\
        & \texttt{[CLS]} Attention & contiguous & -0.0257 & -19.4711 &  0.0000 \\
        &         & Non contiguous & -1.0000 &  -1.0000 & -1.0000 \\
Evidence Inference & Gradient & contiguous & -0.0126 &  -0.5457 &  0.6143 \\
        &         & Non contiguous & -0.0139 &  -0.9352 &  0.4026 \\
        & \texttt{[CLS]} Attention & contiguous & -0.0145 &  -1.4655 &  0.2166 \\
        &         & Non contiguous &  0.0053 &   0.3776 &  0.7249 \\
Movies & Gradient & contiguous & -0.0221 &  -6.4826 &  0.0029 \\
        &         & Non contiguous & -0.0020 &  -0.2684 &  0.8016 \\
        & \texttt{[CLS]} Attention & contiguous & -0.0232 &  -3.1249 &  0.0354 \\
        &         & Non contiguous &  0.0040 &   1.6500 &  0.1743 \\
MultiRC & Gradient & contiguous & -0.0041 &  -1.4573 &  0.2188 \\
        &         & Non contiguous &  0.0066 &   0.8969 &  0.4205 \\
        & \texttt{[CLS]} Attention & contiguous & -0.0038 &  -1.1710 &  0.3066 \\
        &         & Non contiguous &  0.0012 &   0.2832 &  0.7910 \\
\bottomrule
\end{tabular}

    \caption{Comparison of global rationales vs instance level rationale for each dataset, saliency and rationale type combination. The statistical test used was Welch's $t$-test (2-sided). $\Delta$ = (Average F1 score for global) - (average F1 score for instance level) heuristics.}
    \label{tab:global}
\end{table*}

\section{Global Discretization Heuristics} 
\label{app:global}
We train $\rat$-models based on the following global rationales:

\paragraph{Global top-$k$.}
A ratio of tokens to maintain $p$ is determined beforehand
, but instead of taking the top $p\cdot |x_i|$ tokens from each instance, the training set (resp. dev, test) is created by only taking the top $p\cdot \sum_{x_i\in\mathcal{D_{tr}}}|x_i|$ tokens from the entire initial training set (resp. dev, test).
To avoid the possible complete removal of certain instances, we add a further constraint where each instance first secures the top $q<p$-proportion tokens, before filling up the remainder globally.

\paragraph{Global Contig.}
To limit the rationales to a contiguous text span, we first find the maximum-mass segments for the appropriate range of lengths on each instance, then run a greedy algorithm to find per-instance lengths for overall maximal mass:
starting with the minimally-long spans, of total length $L_m$, we perform $\mathcal{B}-L_m$ iterations of finding the next single-token addition in the entire dataset which will lead to the maximum increase in overall weight (each time ranking the marginal gains for each instance $x_i$ of replacing the current $k(x_i)$-length span with the best $k(x_i)+1$-length span in the instance).

In Table \ref{tab:global} we provide average differences between using global vs instance heuristics to extract rationales from our documents, given saliency scores. We also ran a $t$-test to determine if global heuristics provided results significantly different from instance-level methods, finding that they did not.

\end{document}